\documentclass{article}
\usepackage{ijcai24}
\usepackage{CJKutf8}
\usepackage{times}
\usepackage{soul}
\usepackage{url}
\usepackage{enumitem}
\usepackage{xspace}
\usepackage[hidelinks]{hyperref}
\usepackage[utf8]{inputenc}
\usepackage[small]{caption}
\usepackage{graphicx}
\usepackage{amsmath}
\usepackage{amssymb}
\usepackage{amsthm}
\usepackage{booktabs}  
\usepackage{multirow}
\usepackage{multicol}  
\usepackage[switch]{lineno}
\usepackage{subcaption}

\usepackage[table]{xcolor}
\usepackage{diagbox}

\usepackage[ruled,vlined]{algorithm2e}

\def\model{TrajCL\xspace}

\newtheorem{Def}{Definition}

\newtheorem*{Pro*}{Problem Statement}

\title{Towards Robust Trajectory Representations: Isolating Environmental Confounders with Causal Learning}

\author{
Kang Luo$^1$\and
Yuanshao Zhu$^1$\and
Wei Chen$^1$\and
Kun Wang$^2$\and \\
Zhengyang Zhou$^2$\and
Sijie Ruan$^3$\and
Yuxuan Liang$^{1}$\footnote{Corresponding author. Email: yuxliang@outlook.com}
\\
\affiliations
$^1$Hong Kong University of Science and Technology (Guangzhou)\\
$^2$University of Science and Technology of China
~$^3$Beijing Institute of Technology\\
\emails
\{connorluo, yuxliang\}@outlook.com,
yuanshaozhu@hkust-gz.edu.cn,
onedeanxxx@gmail.com,
wk520529@mail.ustc.edu.cn,
zzy0929@ustc.edu.cn,
sjruan@bit.edu.cn
}

\begin{document}

\maketitle

\begin{abstract}

Trajectory modeling refers to characterizing human movement behavior, serving as a pivotal step in understanding mobility patterns.
Nevertheless, existing studies typically ignore the confounding effects of geospatial context, leading to the acquisition of spurious correlations and limited generalization capabilities. 
To bridge this gap, we initially formulate a Structural Causal Model (SCM) to decipher the trajectory representation learning process from a causal perspective. Building upon the SCM, we further present a \underline{Traj}ectory modeling framework (\textit{\model}) based on \underline{C}ausal \underline{L}earning,  which leverages the backdoor adjustment theory
as an intervention tool to eliminate the spurious correlations between geospatial context and trajectories.
Extensive experiments on two real-world datasets verify that \model markedly enhances performance in trajectory classification tasks while showcasing superior generalization and interpretability.

\end{abstract}

\section{Introduction}
Trajectory data has emerged as an indispensable resource for understanding human mobility patterns~\cite{jin2023spatio}. Such data offers invaluable insights into various applications ranging from traffic management to personalized location-based services~\cite{dai2015personalized,chen2022mutual}. As a result, the modeling of this data is the cornerstone for transforming raw location information into mobility intelligence, thereby supporting various spatial-temporal applications, e.g., travel mode detection~\cite{zheng2008learning,zhu2021semi}, next location prediction~\cite{yin2016adapting,lin2021pre}, and travel time estimation~\cite{zheng2015trajectory}.

\emph{Trajectory representation learning} involves extracting useful, generalizable, and concise representations from the sequential data points of a human trajectory.
Intuitively, its functionality extends to discerning the intrinsic motion properties inherent in trajectories.
This pursuit is typically realized by deep sequential models, such as Recurrent Neural Networks \cite{wu2017modeling,liu2019spatio} and Transformers \cite{liang2022trajformer,chang2023contrastive}, which effectively capture trajectory dynamics by encoding temporal intervals and spatio-temporal correlations \cite{liu2017end,qin2019toward,liu2019spatio,liang2021modeling}. 
Furthermore, considering the geospatial context associated with trajectories, e.g., point of interests, road networks, recent endeavors are devoted to mining valuable insights from this auxiliary information~\cite{guo2020context,ferrero2020mastermovelets,pugliese2023summarizing,jiang2023self,wang2024cola}. 

\begin{figure}[t]
  \centering
  \includegraphics[width=0.9\linewidth]{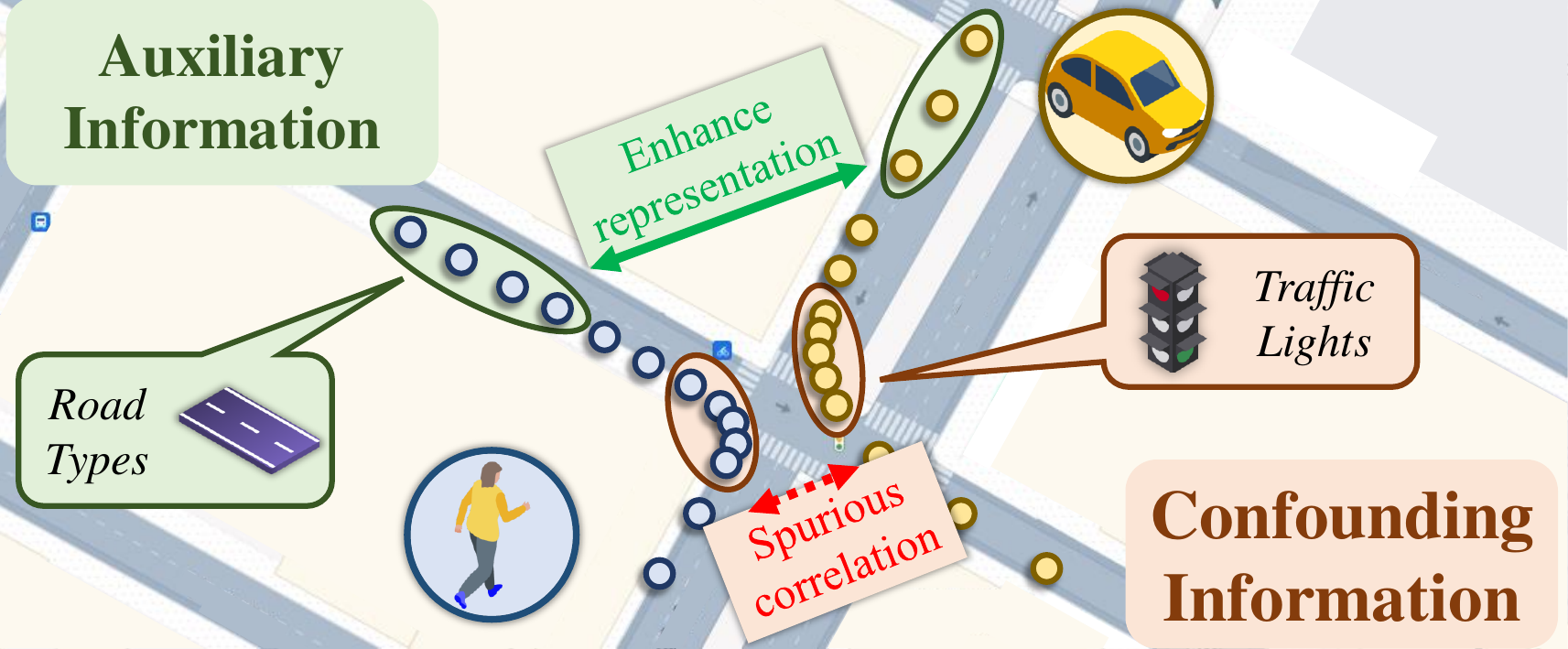}
  \vspace{-2mm}
  \caption{The impacts of geospatial context on trajectory modeling.}
  \vspace{-3mm}
  \label{fig:intro}
\end{figure}

While the integration of the geospatial context has the potential to enhance trajectory representations to a certain degree, it concurrently introduces a \emph{confounding factor} into the learning process. 
This confounder poses the risk of our algorithm learning spurious correlations within the training data, leading to a degradation in performance and a compromised ability to generalize. 
In other words, the model is vulnerable to overfitting to specific environmental conditions. 
To elucidate this concern, consider an example in Figure \ref{fig:intro}. Vehicles frequently come to a halt in congested areas or traffic lights, exhibiting trajectory patterns (e.g., low speed) akin to those of pedestrians.
In this scenario, there exists an increasing risk of the model displaying a pronounced inclination towards recognizing pedestrian patterns in congested areas. This, in turn, could lead to the \emph{spurious correlation} (i.e., unwarranted association) of geospatial context with trajectory patterns.

In this paper, our target is to mitigate the impact of these confounding factors induced by the geospaital context, so as to extract \textit{robust} and \textit{domain-invariant} representations from human trajectories. Primarily, we present a \emph{Structural Causal Model} (SCM) to deepen our comprehension of the trajectory representation learning process. From a causal perspective, the SCM elucidates the relationship among the environment (i.e., geospatial context\footnote{we use geospatial context and environment interchangeably.}), the trajectory, and the representation outcome. The environment serves as a confounding factor that establishes a backdoor path between the input trajectory data and the resulting representation. 

Building upon the SCM, we present a novel \underline{c}ausal \underline{l}earning framework called \textbf{\model} for learning robust \underline{traj}ectory representations. \model utilizes the \emph{backdoor adjustment} theory as an intervention tool to eliminate the spurious correlations between environment and trajectories, which is implemented by two key steps. Firstly, we design an environmental alignment module that leverages geospatial context to guide the encoders in disentangling causal and confounding representations. Secondly, we elaborately introduce a causal learning module to effectively accomplish causal intervention at the representation level, resulting in robust representations that exhibit strong generalization capabilities, particularly in few-shot learning or imbalanced sample learning scenarios.

Our major contributions can be summarized as follows:
\begin{itemize}[leftmargin=*]

\vspace{-0.1em}
\item \emph{A causal lens for trajectory data.} We propose a structural causal model to unravel the inherent rationale behind the learning process of trajectories. Based on this causal lens, a novel framework termed TrajCL is presented to enhance the robustness of trajectory representations.

\vspace{-0.1em}
\item \emph{Backdoor adjustment for isolating confounders.} Exploiting the backdoor adjustment theory, we initially design an environment alignment module to disentangle the causal and confounding elements from input data, and subsequently leverage causal intervention to learn robust representations.

\vspace{-0.1em}
\item \emph{Extensive empirical studies.} We conduct thorough experiments on the trajectory classification task over two mobility datasets. The results affirm that our framework enhances trajectory modeling in a plug-and-play manner, and possesses superior generalization ability and interpretability.

\end{itemize}

\section{Related Work}

\par \noindent \textbf{Trajectory Modeling.}
Extensive research has been dedicated to trajectory modeling, aiming to uncover human mobility patterns. Early heuristic-based approaches only considered limited temporal or spatial dimensions. ~\cite{lee2008traclass} initially employed a spatial grid method, merging trajectory points within single grid units, exploring the spatial characteristics of trajectory substructures. Building upon this,~\cite{zheng2008learning} and~\cite{dodge2009revealing} respectively extracted local and global features from subgrids and trajectory points for travel mode classification. Moreover,~\cite{xiao2017identifying} incorporated semantic information, such as road networks, to classify vehicle trajectories. However, such works heavily rely on handcrafted features and make excessive parameter assumptions. Recently, studies in trajectory modeling have focused on deep representation learning, benefiting from its superior modeling capabilities in sequential data~\cite{chen2024deep}. 
~\cite{liu2017end} and~\cite{jiang2017trajectorynet} utilized two common RNN architectures to capture high-order movement patterns. 
After that,~\cite{liu2019spatio} and~\cite{liang2021modeling} introduced segment-wise convolutional weighting mechanisms and neural differential equations to enforce RNNs with the capability of continuous time updates, addressing the modeling of trajectory's continuous temporal characteristics. 
~\cite{han2021graph} and~\cite{yao2022trajgat} integrated spatial features from road networks by graph neural networks, enabling the capture of long-term dependencies in trajectories.
Furthermore, TrajFormer~\cite{liang2022trajformer} adapted an advanced transformer architecture to balance speed and accuracy in trajectory modeling. 
Nevertheless, due to the inherent noise in raw trajectory data and environmental biases, obtaining robust trajectory representations with causal invariance remains challenging.

\par \noindent \textbf{Causal Inference.} Traditional causal inference aims to study how to learn a causal model that works under different distributions, encompassing causal mechanisms, and further employs causal models for intervention or counterfactual inference~\cite{pearl2009causal}. However, real-world observations often do not begin with basic inference units (random variables connected in a causal graph) but rather with high-dimensional raw data. Therefore, causal representation learning~\cite{scholkopf2021toward} seeks to integrate deep learning and causal inference, widely explored in various fields such as computer vision~\cite{lippe2022citris}, recommendation systems~\cite{wang2023causal}, graph data mining~\cite{sui2022causal} and so on. Nevertheless, research in trajectory modeling is still in its early stages, with only a limited amount of literature addressing spatio-temporal data mining.~\cite{li2023transferable} investigated distribution changes in time series from a causal perspective, but they neglected to decipher spatial factors. Subsequently,~\cite{deng2023spatio} constructed a causal graph to describe traffic prediction and analyze the causal relationships between spatio-temporal features and outcomes. Similarly, ~\cite{xia2023deciphering} applied it to spatio-temporal graph forecasting, mitigating confounding effects in the temporal and spatial domains using causal inference. Still and all, they are all graph-based models considering causal effects and cannot adapt to trajectory data as spatio-temporal sequences. In this study, we adopt a causal perspective to investigate trajectory modeling and use causal techniques to mitigate confounding factors in the environment.
\section{A Causal View on Trajectory Modeling}
\subsection{Formulation}

\begin{Def}[Trajectory]
We denote a trajectory by $X=\{p_i \mid i=1, 2, \ldots, n\}$ with a sequence of spatio-temporal points $p_i$ recorded in chronological order. Each point ${p}=\left({Lon},{Lat},t\right)$ is a longitude, latitude, and timestamp triplet.
\end{Def}

\begin{Def}[Geospatial Context]
The geospatial context is the set $E$ of environment information, where $e_i \in \mathbb{R}^m$ represents the attributes of the surrounding environment related to the point $p_i$, and $m$ denotes the number of attributes.
\end{Def}

\begin{Pro*}[Trajectory Modeling]
Our target is to learn robust and high-quality trajectory representation $H$ via supervised signals to support various downstream tasks. In our task, we set it up for travel mode identification.
\end{Pro*}

\begin{figure}[!t]
  \begin{subfigure}[b]{0.32\columnwidth}
    \includegraphics[width=\linewidth]{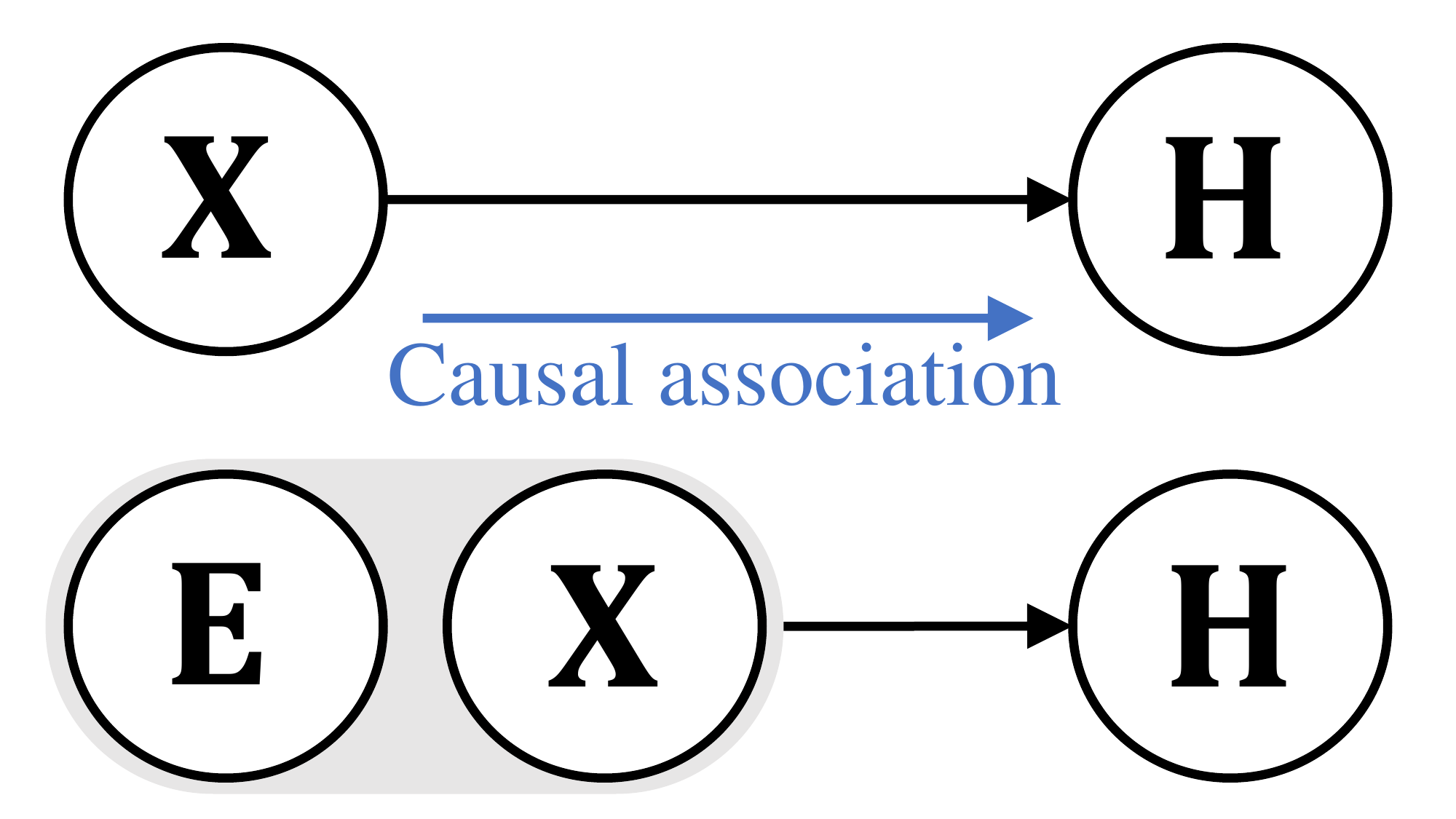}
    \caption{}
    \label{fig-scm1}
  \end{subfigure}
  \begin{subfigure}[b]{0.32\columnwidth}
    \includegraphics[width=\linewidth]{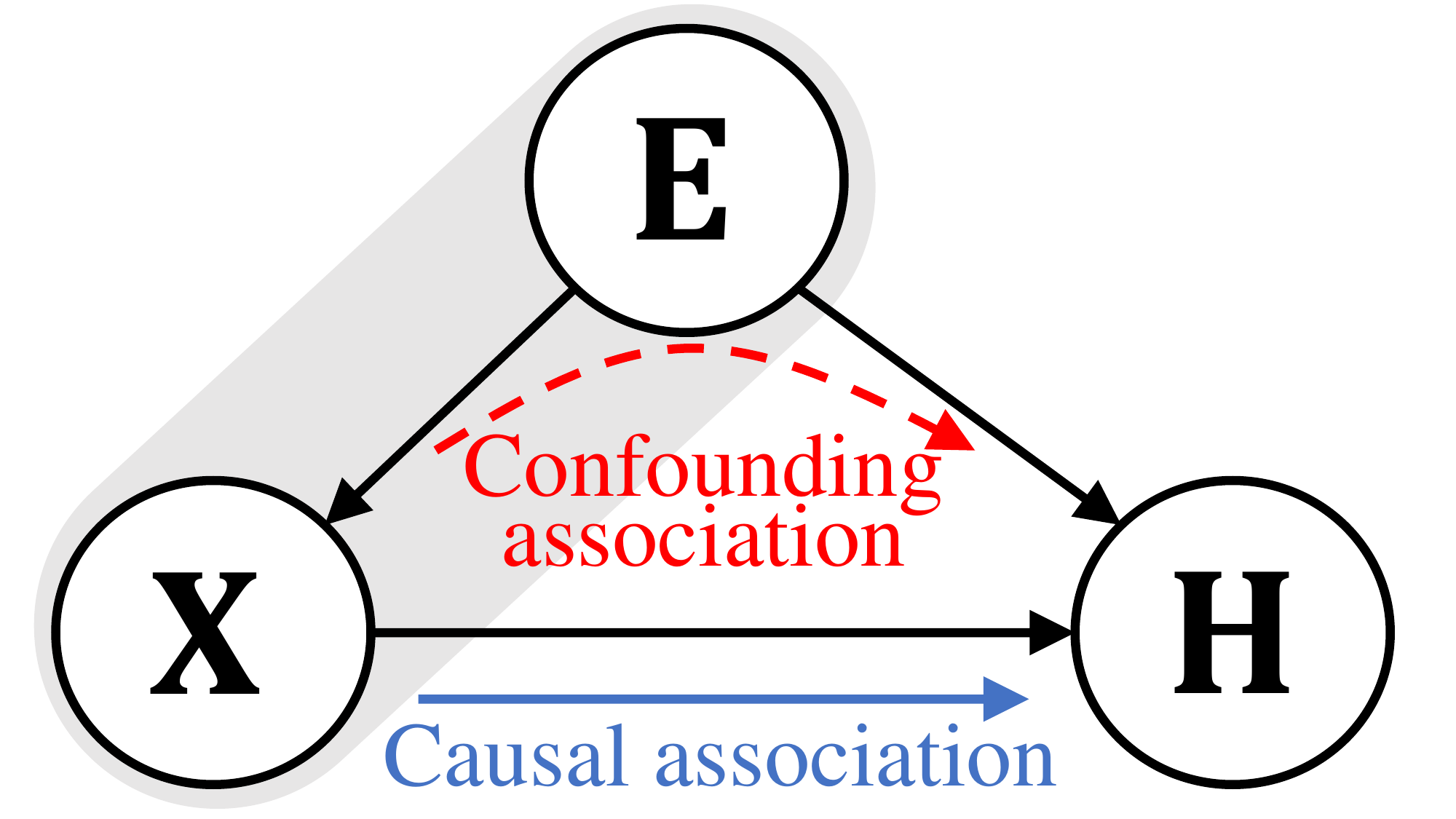}
    \caption{}
    \label{fig-scm2}
  \end{subfigure}
  \begin{subfigure}[b]{0.32\columnwidth}
    \includegraphics[width=\linewidth]{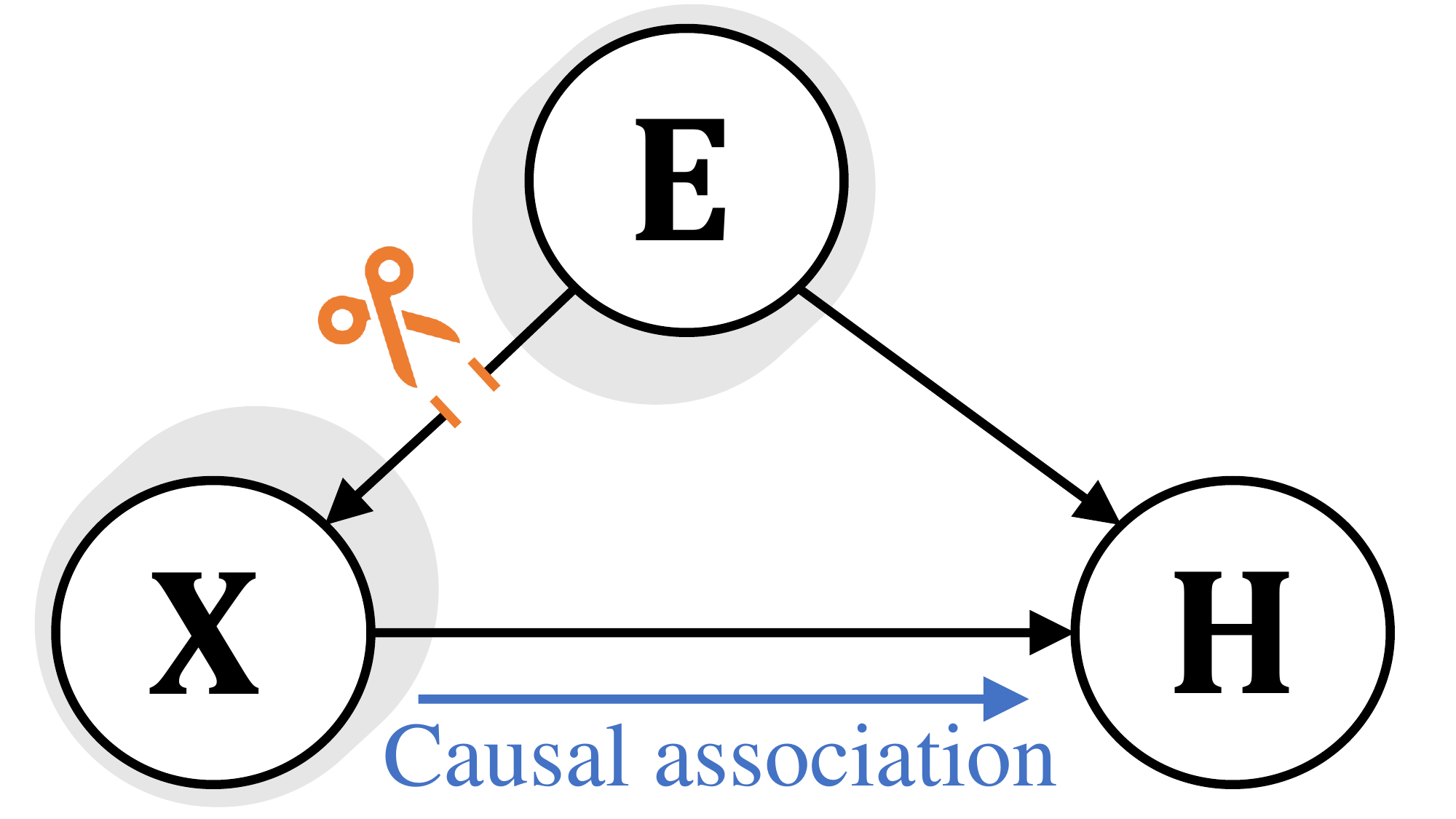}
    \caption{}
    \label{fig-scm3}
  \end{subfigure}
  \vspace{-2mm}
  \caption{SCMs of trajectory modeling. The SCM (a) in traditional perspective; (b) under causal view; (c) after back-door adjustment.}
  \label{fig-scm}
  \vspace{-2mm}
\end{figure}
\subsection{Structural Causal Model}\label{sec:scm}
Formally, we build a Structural Causal Model (SCM) to analyze the causality in trajectory modeling (Fig.~\ref{fig-scm}). It comprises three variables: trajectory data $X$, environment $E$, and trajectory representation $H$, with arrows denoting causal relationships. The SCM provides the following elucidation:

\begin{itemize}[leftmargin=*]
\item $X{\rightarrow}H$.
Trajectories encompass crucial movement characteristics such as speed, acceleration, and direction. Variations in these fundamental features give rise to discernible trajectory representations. Intuitively, trajectory data can be modeled using internal sequential patterns to learn representations: $H=\mathcal{F}\left(\theta ; X\right)$
,where $\mathcal{F}$ denotes the parameterized modeling function with learnable parameters $\theta$.

\item $E{\rightarrow}X$ and $E{\rightarrow}H$.
Trajectory data is profoundly impacted by environmental factors. An illustrative example of this influence is the rarity of pedestrian trajectories along highways. Moreover, the dynamic nature of the environment contributes to alterations in the distribution of latent trajectory representations.  Notably, distinctive motion tendencies emerge among individuals and vehicles in response to environmental cues such as traffic signals.  In light of these environmental factors, the joint modeling of trajectories and environment can be expressed as $H = \mathcal{F} (\theta;~X,~E).$
\end{itemize}

However, upon meticulous examination of the SCM, we identify a \emph{backdoor path} between $X$ and $H$, i.e., $X{\leftarrow}E{\rightarrow}H$, where environmental information acts as a confounding factor for both trajectory data and trajectory representation. This introduces a confounding factor set $\mathcal{E}$, signifying a confounding association between environment and trajectory representation (Fig.~\ref{fig-scm2}), distinct from the causal association from $X$ to $H$ (Fig.~\ref{fig-scm1}). Consequently, in the practical process of jointly learning trajectory-environment representations, this strong correlation comprises a blend of confounding and causal associations. Therefore, we advocate for employing causal intervention techniques on variable $X$ to mitigate the confounding effects of variable $E$ (Fig.~\ref{fig-scm3}), facilitating the acquisition of robust representations from biased trajectory data.

\subsection{Backdoor Adjustment}
Based on the aforementioned causal analysis, our approach to learning trajectory representations involves eliminating backdoor paths rather than modeling the association $P(H|X)$ in Fig.~\ref{fig-scm1}. We employ a powerful tool known as backdoor adjustment, rooted in causal theory~\cite{pearl2009causal}, allowing us to block the backdoor path by estimating $P(H|do(X))$, where $do(\cdot)$ denotes the \textit{do-calculus}. Specifically, we have
\begin{equation}
\label{eq:backdoor}
\resizebox{.91\linewidth}{!}{$
\begin{split}
P(H|do(X)) &={\sum\nolimits_{e\in\mathcal{E}}}{P(H|do(X),e)P(e|do(X))} \\
& =\sum\nolimits_{e\in\mathcal{E}} P(H|do(X),e)P(e) \\
& =\sum\nolimits_{e\in\mathcal{E}} P(H|X, e)P(e),
\end{split}
$}
\end{equation}
where $\mathcal{E}$ represents the set of environment confounding factors. First, we can redefine $P(H|do(X))$ via Bayes' theorem. Then, due to the independence of variables $H$ and $X$ under intervention, resulting in $P(e|do(X))=P(e)$. Simultaneously, the response of $H$ to $E$ and $X$ is unrelated to the causal association between $H$ and $X$, allowing us to equate the conditional probabilities 
$P(H|do(X), e)=P(H|X, e)$.

Nonetheless, the latent confounding factors denoted as $\mathcal{E}$ prove elusive and challenging to directly observe~\cite{xia2023deciphering}. Manipulating extensive trajectory data also poses additional complexities, necessitating adjustment at the representation level through learned strategies. To this end, we introduce two new modules in the following section to implement Eq.~\ref{eq:backdoor} for trajectory representation learning.

\section{Model Implementation}\label{methods}

\begin{figure*}[t!]
    \vspace{-4mm}
    \centering
    \includegraphics[width=0.9\textwidth]{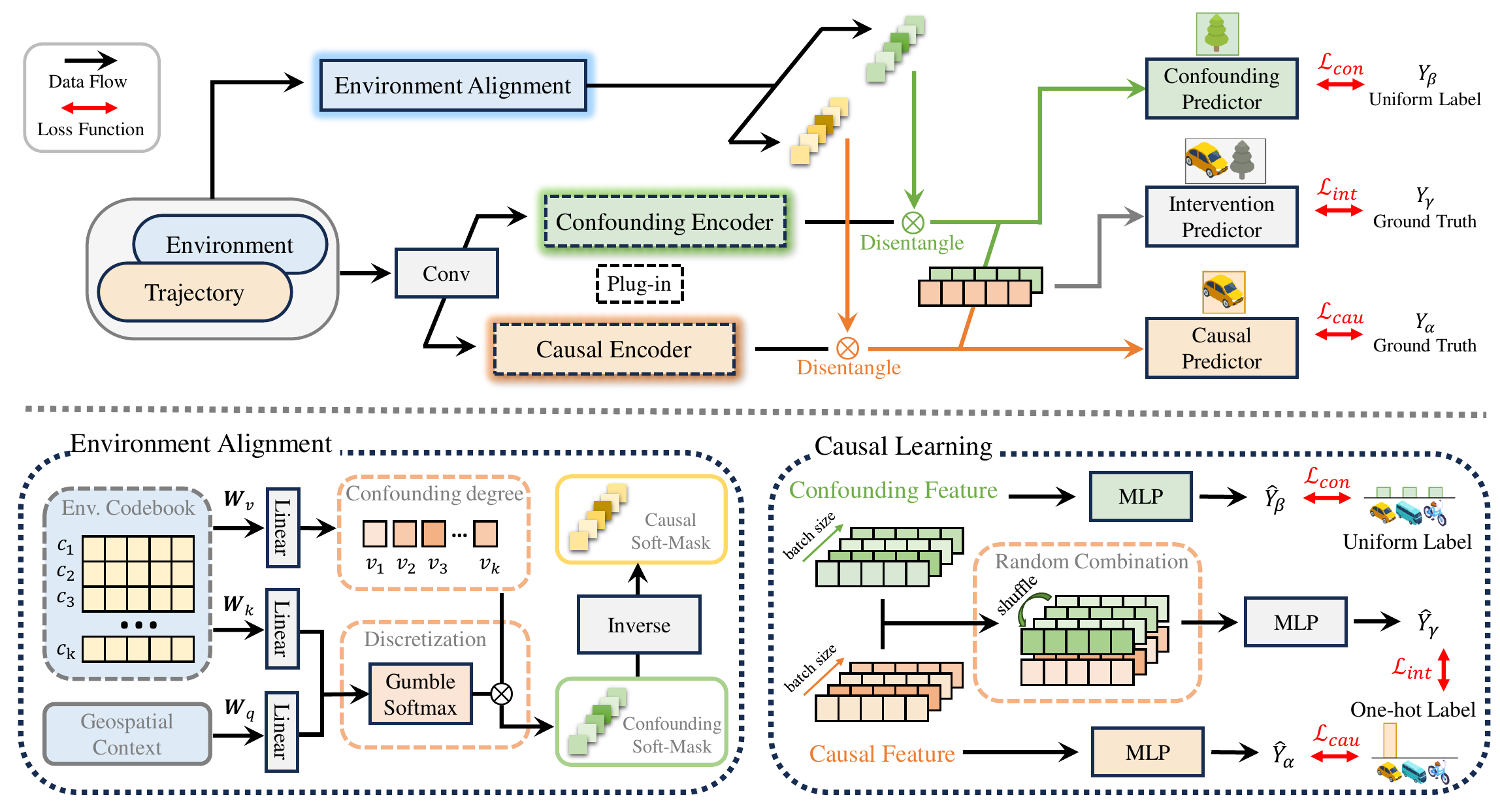}
    \vspace{-3mm}
    \caption{The architecture of the proposed \model framework. Env: Environment.}
    \label{fig-framework}
    \vspace{-4mm}
\end{figure*}

In this section, we present the details of our TrajCL framework (see Fig.~\ref{fig-framework}). We first provide an \textit{overview of the \model framework}, integrating causal techniques into the trajectory modeling. Then, we delve into the specifics of two crucial components: \textit{environmental alignment module} and \textit{causal learning module}. These tailored modules are designed to be compatible with any advanced trajectory model, allowing for efficient enhancement from a causal perspective.

\subsection{Framework Overview}

Traditional trajectory learning methods mostly involve convolutional layers to extract local features from input $X$ at the point level.
A deep sequential model is subsequently employed to capture spatio-temporal patterns at the trajectory level. 
Notably, the geospatial context $E$ corresponding to each trajectory point is regarded as auxiliary information incorporated into the encoding process. 
For clarity, we use the term encoder $\mathcal{F}$ to represent the aforementioned sequential encoding process in this paper. 

As analyzed in section~\ref{sec:scm}, the simple introduction of environmental information can lead to confounding factors.
Therefore, we extend the representation modeling strategy from a causal perspective. 
The sequence pattern encoder $\mathcal{F}$ is duplicated into two counterparts, namely the causal encoder $\mathcal{F}_{\alpha}$ and the confounding encoder $\mathcal{F}_{\beta}$, maintaining the identical architecture but without parameter sharing. The formalized process is as follows:
\begin{equation}
\begin{split}
    H_{\alpha} = \mathcal{F}_{\alpha} (\theta_1;~X,~E),~~
    H_{\beta} = \mathcal{F}_{\beta} (\theta_2;~X,~E).
\end{split}
\end{equation}
$\mathcal{F}_{\alpha}$ and $\mathcal{F}_{\beta}$ denotes the encoder, which can be shifted to advanced trajectory representation models, e.g., GRU. 
$H_{\alpha}$ and  $H_{\beta}$ denote the the embedding of original sequence.

Guided by a specially developed \textit{environmental alignment module}, the features output by the two encoders are separately disentangled. 
Specifically, $H_{\alpha}$ and  $H_{\beta}$ are fed into two streams. 
One stream is dedicated to extracting invariant representations of the former, termed causal features (represented by the orange dashed line in Fig.~\ref{fig-framework}). 
The other stream aims to extract the confounding effects caused by the environment, termed confounding features (as indicated by the green dashed line in Fig.~\ref{fig-framework}). 
Finally, a carefully designed \textit{causal learning module} intervenes at the representation level, jointly with downstream tasks, to differentiate causal and confounding features and achieve robust representation acquisition.

\subsection{Environment Alignment Module}\label{sec:env_module}
The goal of this module is to learn two soft-masks for confounding awareness and causal awareness, guiding the two streams to disentangle causal and confounding features from the input data. 
It comprises two sub-modules: a \textit{cross attention component} and a \textit{disentanglement allocation component}.

\subsubsection{Cross Attention Component}

From an intuitive view, various types of environments inherently function as confounding factors to differing degrees.
Therefore, we first create a learnable environment codebook $C=\{c_1,c_2,...,c_k\}$, where $C\in\mathbb{R}^{k \times d}$ represents the prototype embedding of urban environments.  
Subsequently, each prototype is fed into a linear layer to encode its individual confounding degree:
\begin{equation}
    V={W}_{v}C+b_{v},
\end{equation}
where ${W}_v\in\mathbb{R}^{d\times1}$ and $b_v$ are learnable parameters, and $V\in\mathbb{R}^{k\times1}$ denotes the confounding degrees corresponding to the environment codebook, i.e., $V=\{v_1,v_2,...,v_k\}$.

Meanwhile, we project the environment codebook and the geospatial context $E\in\mathbb{R}^{n \times m}$ into the same latent space:
\begin{equation}
\begin{split}
    Q={W}_{q}E+b_{q},
    ~~K={W}_{k}C+b_{k},
\end{split}
\end{equation}
where ${W}_q\in\mathbb{R}^{m\times d}$ and ${W}_k\in\mathbb{R}^{d\times d}$ are learnable parameter matrices for geospatial context and the codebook, respectively. $b_{q}$ and $b_{k}$ are bias vectors.

Next, we apply a cross-attention mechanism to assess the similarity scores among all prototypes in the environment codebook $Q$ with geospatial context $K$. Specifically:
\begin{equation}
    M_{\alpha}=\operatorname{Gumber-Softmax}({{Q}\cdot{K}^{\mathsf{T}}/{\sqrt{d}}})\cdot{V},
\end{equation}
where $M_{\alpha}\in\mathbb{R}^{n\times{1}}$ is the adjusted confounding intensity, serving as a confounding soft-mask, and its complement is the causal soft-mask $M_{\beta}=1-M_{\alpha}$. Notably, allocating all prototypes to the environment contradicts our intuition. We hence employ the Gumbel-Softmax~\cite{jang2016categorical} to select the most similar environment, expressed as:
\begin{equation}
    \mathbf{s}_{i}=\frac{e^{\left.\left(\log \left(\pi_{i}\right)+\mathcal{G}_{i}\right) / \tau\right)}}{\sum_{j=1}^{k} e^{\left.\left(\log \left(\pi_{j}\right)+\mathcal{G}_{j}\right) / \tau\right)}},
\end{equation}
where $\mathbf{s}$ is the $n$-dimensional vector, $\pi$ are class probabilities, $\mathcal{G}\sim\operatorname{Gumbel}(0,1)$ are i.i.d samples drawn from the Gumbel distribution, and $\tau$ is the temperature. Due to orthogonal properties, it enforces discretized similarity learning.

\subsubsection{Disentanglement Allocation Component}

After completing the cross-attention process, we apply the two masks $M_{\alpha}$ and $M_{\beta}$ to the representations from causal and confounding encoders, allocating weights to each spatio-temporal point. 
This process enables the encoders $\mathcal{F}_{\alpha}$ and $\mathcal{F}_{\beta}$ to learn disentangled representations.
Moreover, for both branches, we perform average pooling operations along the time step dimension:
\begin{equation}
\begin{split}
Z_{\alpha}= \operatorname{Pooler}(H_{\alpha} \odot {M}_{\alpha}), 
~~Z_{\beta}= \operatorname{Pooler}(H_{\beta} \odot {M}_{\beta}),
\end{split}
\end{equation}
where $\odot$ denotes the Hadamard product operation. $Z_{\alpha}$ and $Z_{\beta}$ are trajectory-level causal and confounding features.

\subsection{Causal Learning Module}

We further implement diverse strategies for parameterizing backdoor adjustment. This involves a \textit{disentangle learning strategy} that combines downstream tasks to distinguish causality from confounding, an \textit{intervention learning strategy} to eliminate confounding, and a final \textit{optimization} process.

\subsubsection{Disentangle Learning Strategy}

Given our objective of travel mode identification, we choose to employ category classifiers via a multilayer perceptron (MLP) across two branches. The purpose of causal features is to estimate invariant characteristics within trajectories, leading to the causal branch output being classified with actual travel mode labels. In contrast, confounding features aim to represent information unrelated to the intrinsic patterns of trajectories. Consequently, the output from the confounding branch is distanced from travel mode representations, approaching the average label across all categories. Formally, we define two classification losses:
\begin{equation}
\begin{split}
    \hat{Y}_{\alpha}=\operatorname{MLP}(Z_{\alpha}),~~\hat{Y}_{\beta}=\operatorname{MLP}(Z_{\beta}), \\
    \mathcal{L}_{cau} = \mathcal{H}(Y_{\alpha}, \hat{Y}_{\alpha}),~~\mathcal{L}_{con} = \mathcal{H}(Y_{\beta}, \hat{Y}_{\beta}),
\end{split}
\end{equation}
where ${Y}_{\alpha}$ represents the ground-truth travel mode label, ${Y}_{\beta}$ is assigned using a uniform distribution as the ground-truth label, $\mathcal{H}(\cdot,\cdot)$ refers to the conventional cross-entropy loss.

\subsubsection{Intervention Learning Strategy}

To achieve invariant predictions in a dynamic environment, the elimination of confounding factors emerges as a pivotal endeavor. To this end, we perform interventions by hierarchically manipulating the confounding features and randomly combining them with causal features to achieve backdoor adjustment in Equation~\ref{eq:backdoor}. Specifically, we merge the causal features of one input with the confounding features of another randomly chosen input from the same batch as the intervention features. Given the inherent characteristic that confounders do not exert influence on the final outcome, the prediction results will align well with the actual travel mode as:
\begin{equation}
\begin{split}
    \hat{Y}_{\gamma}=\operatorname{MLP}(Z_{\alpha}+Z^{'}_{\beta}),~~
    \mathcal{L}_{int} = \mathcal{H}(Y_{\gamma}, \hat{Y}_{\gamma}),
\end{split}
\end{equation}
where ${Y}_{\gamma}$ is also the travel mode ground-truth label.

\subsubsection{Optimization}

The objective function of \model can be defined as the sum of the above three losses:
\begin{equation}
\mathcal{L} = \lambda\mathcal{L}_{cau} + \varphi\mathcal{L}_{con} +  \eta\mathcal{L}_{int},
\end{equation}
where $\lambda$, $\varphi$ and $\eta$ are weight hyperparameter.
Through backpropagation optimization, we can force the causal representation to be invariant and stable, the confounder representation to be independent, and the intervention representation to be consistent. Finally, we use the causal representation as a robust trajectory representation.

\section{Experiments}
In this section, we evaluate our \model based on an extensively studied trajectory classification task \cite{liang2022trajformer} to answer the following research questions (RQs): 
\begin{itemize}[leftmargin=*]
  \item \textbf{RQ1:}~Does \model really help improve travel mode classification performance, and to what extent?
  \item \textbf{RQ2:}~How robust is \model under varying conditions?
  \item \textbf{RQ3:}~How do the key components and the hyperparameters of \model affect the performance?
  \item \textbf{RQ4:}~Does \model accurately and efficiently yield explanations, and how intuitively understandable is it?
\end{itemize}

\subsection{Experimental Settings}  

\par \noindent \textbf{Datasets}

\noindent We conduct extensive experiments on two publicly real-world datasets GeoLife~\cite{zheng2009mining} and Grab-Posisi~\cite{huang2019grab}: GeoLife consists of human trajectories collected by 182 users (73 of whom are labeled with their travel mode) in Beijing from April 2007 to August 2012. Following \cite{liu2019spatio}, we divide trajectories into segments and classify them into four typical travel modes: walking, bus, bike, and driving. Grab-Posisi is a dataset of delivery trajectories collected by Grab, a Southeast Asian ride-hailing company. We follow \cite{liang2021modeling} to pick data over a two-week period in Jakarta, with travel modes including car and motorcycle. In addition, we grid the city into cells of $200  \times 200~\text{square meters}$ for each dataset. To capture environmental factors, we then retrieve 24 fixed geospatial features (such as the number of traffic lights, crossings, and residential areas) for each grid from OpenStreetMap.com as indicators of environmental conditions. These environmental variables are then assigned to each GPS point.
After preprocessing, we collect 26,509 trajectories for GeoLife and 507,522 for Grab-Posisi. Each trajectory comprises 20 to 50 GPS points. These trajectories are subsequently partitioned in an 8:1:1 ratio for training, validation, and test data.

\par \noindent \textbf{Baselines \& Evaluation Metrics}

\noindent To validate the effectiveness and robustness of our proposed \model, we implement our framework with five representative baseline models for trajectory modeling, including vanilla GRU, BiLSTM~\cite{liu2017end}, GRU-D~\cite{che2018recurrent}, STGN~\cite{zhao2020go}, and TrajFormer~\cite{liang2022trajformer}. We refine them for optimal performance using the parameter settings suggested in their papers. 
Then, we follow~\cite{liu2017end,liang2022trajformer} to employ the classification accuracy (Acc.) to evaluate model performance. Each model and setting is run on every dataset thrice, and the mean accuracy is reported. Besides, the symbol $\Delta$ denotes accuracy change.

\par \noindent \textbf{Implementation Details \& Hyperparameters}

\noindent  Our model uses the Adam optimizer with the initial learning rate set to $0.001$, reduced by 0.1 every 30 epochs. 
To avoid overfitting, we employ an early stopping mechanism with a patience of 20 epochs. 
The batch sizes for the GeoLife and Grab-Posisi datasets are 256 and 512, respectively. 
The default embedding dimensions are set to 64. 
For the predictor, we apply a 2-layer MLP uniformly. 
To initially merge the trajectory and geospatial inputs, we employ two $3\times1$ convolutional layers. 
The weight parameters $\lambda$, $\varphi$, and $\eta$ of the loss are 1, 0.5, and 0.5, respectively.

\subsection{Overall Performance (RQ1)}\label{sec:rq1}
Table~\ref{tab:comp} depicts the overall performance under different model settings on the two datasets.
Specifically, we report the performance of the original baseline (Base), the results with incorporating environmental information (+ Env), and applied with the \model framework (+ \model).
The experimental results demonstrate a clear overall trend: incorporating environmental information significantly enhances the performance, which underscores the importance of considering environmental influences in trajectory modeling.
However, incorporating environmental information into trajectory modeling may simultaneously introduce confounding factors that negatively affect the modeling process.
This phenomenon is further validated by applying \model framework, where we can observe more improvements after using \model for causal intervention.
The above results strongly suggest that there are indeed confounding effects within environments, and emphasize the effectiveness and necessity of our proposed \model.

In addition, the environmental factors and the \model present different influences across the two datasets.
In the GeoLife dataset, the overall improvement of incorporating environmental factors is less significant compared to the Grab-Posisi dataset (average +4.37\% in Geolife, and +10.15\% in Grab-Posisi).
This can be attributed to the high frequency and diversity of confounders in the Geolife dataset, which partially undermines the contribution of environmental factors.
Comparably, the Grab-Posisi dataset is limited to car and motorcycle travel modes with less surrounding noise.
Therefore, when the \model intervention is applied to mitigate these confounding factors, the performance improvement is more substantial in the GeoLife dataset (average +1.41\% in Geolife, and +0.96\% in Grab-Posisi). 
This difference highlights the effectiveness of the \model intervention, particularly in scenarios with diverse and complex environmental confounders.

\begin{table}[t]
\small
\setlength{\tabcolsep}{2.5pt}

\centering
\scalebox{1}
{
\begin{tabular}{l ccc ccc} 
\toprule
\multirow{2}{*}{\diagbox[width=6em]{Model}{Dataset}} & \multicolumn{3}{c}{GeoLife} & \multicolumn{3}{c}{Grab-Posisi} \\
\cmidrule(lr){2-4}\cmidrule(lr){5-7}
& Base & +~Env  &  +~TrajCL  & Base & +~Env  & +~TrajCL\\ 
\midrule

\multirow{2}{*}{GRU} 
& 69.15 & 75.78 & 77.74 & 63.46 & 78.76 & 79.47\\
& \cellcolor{gray!15}-- & \cellcolor{gray!15}+6.63 & \cellcolor{gray!15}+8.59 & \cellcolor{gray!15}-- & \cellcolor{gray!15}+15.30 & \cellcolor{gray!15}+16.01  \\
\cmidrule(lr){1-7}
\multirow{2}{*}{BiLSTM} 
& 70.20 & 76.58 & 78.02 & 69.07 & 79.81 & 80.70\\
&\cellcolor{gray!15}-- &\cellcolor{gray!15}+6.38 &\cellcolor{gray!15}+7.82 &\cellcolor{gray!15}-- &\cellcolor{gray!15}+10.74 &\cellcolor{gray!15}+11.63  \\
\cmidrule(lr){1-7}
\multirow{2}{*}{GRU-D} 
& 72.48 & 77.37 & 78.62 & 71.78 & 79.17 & 79.69\\
&\cellcolor{gray!15}-- &\cellcolor{gray!15}+4.89 &\cellcolor{gray!15}+6.14 &\cellcolor{gray!15}-- &\cellcolor{gray!15}+7.39 &\cellcolor{gray!15}+7.91  \\
\cmidrule(lr){1-7}
\multirow{2}{*}{STGN} 
& 76.75 & 79.03 & 80.27 & 72.94 & 81.24 & 82.74\\
&\cellcolor{gray!15}-- &\cellcolor{gray!15}+2.28 &\cellcolor{gray!15}+3.52 &\cellcolor{gray!15}-- &\cellcolor{gray!15}+8.30 &\cellcolor{gray!15}+9.80 \\
\cmidrule(lr){1-7}
\multirow{2}{*}{TrajFormer} 
& 78.40 & 80.05 & 81.22 & 76.30 & 85.34 & 86.53\\
&\cellcolor{gray!15}-- &\cellcolor{gray!15}+1.65 &\cellcolor{gray!15}+2.82 &\cellcolor{gray!15}-- &\cellcolor{gray!15}+9.04 &\cellcolor{gray!15}+10.23 \\
    \bottomrule
    \end{tabular}
}
    \vspace{-2mm}
\caption{Performance comparison with different backbones. Acc. report percentage (\%) with $5$ runs average. }
\vspace{-4mm}
\label{tab:comp}
\end{table}

\subsection{Robustness Test (RQ2)}
In this section, we analyze the robustness of \model in the trajectory modeling task with two different scenarios.

\par \noindent \textbf{Few-shot Learning.}
For few-shot learning, we divide the original data set into subsets with ratio $[0.1,~0.2,~0.5]$, and implement on three state-of-the-art models. The results in Table~\ref{tab:subset} show that the smaller the subset, the more significant the improvement provided by \model.
This trend is evident across different models and datasets, demonstrating its high robustness and adaptability to few-shot learning. Notably, the better-performing models (e.g., TrajFormer) exhibit larger \model boosting effects. This suggests that these models, despite having high baseline performance, also \textit{overfit the geospatial context} and introduce more confounding factors when modeling trajectories. 
Similarly, the improvement in the Grab-Posisi dataset is not as significant as in the GeoLife dataset. 
The observation is consistent with our findings in Section~\ref{sec:rq1}, reinforcing the conclusion that the \model is effective in more complex and diverse environments.

\begin{table}[t]
\setlength{\tabcolsep}{1.5pt}
\small
\begin{tabular}{cc cc cc cc cc cc cc cc cc} 
\toprule
\multirow{2}{*}{\diagbox[width=6em]{Dataset}{Model}} & \multicolumn{2}{c}{GRU-D} & \multicolumn{2}{c}{STGN} & \multicolumn{2}{c}{TrajFormer} \\
\cmidrule(lr){2-3}\cmidrule(lr){4-5}\cmidrule(lr){6-7}
& +~Env & +~TrajCL & +~Env & +~TrajCL & +~Env &\cellcolor{gray!15}+~TrajCL \\
\midrule
GL-0.5   & 73.97 & +0.74 & 75.80 & +0.98 & 76.33 &\cellcolor{gray!15}+2.51 \\
GL-0.2   & 70.32 & +0.94 & 71.40 & +1.44 & 72.12 &\cellcolor{gray!15}+2.33 \\
\rowcolor{gray!15}
GL-0.1   & 65.98 & +1.44 & 68.61 & +1.37 & 69.26 &\cellcolor{gray!30}+2.94 \\
\cmidrule(lr){1-7}
GP-0.5& 75.82 & +0.85 & 78.70 & +0.43 & 79.97 &\cellcolor{gray!15}+1.15 \\
GP-0.2& 72.19 & +0.93 & 73.29 & +0.58 & 74.11 &\cellcolor{gray!15}+1.08 \\
\rowcolor{gray!15}
GP-0.1& 69.36 & +1.09 & 70.51 & +1.47 & 71.38 &\cellcolor{gray!30}+1.53 \\
\bottomrule
\end{tabular}
\vspace{-2mm}
\caption{Performance in the few-shot learning setting. Accuracy is reported by percentage (\%). GL: GeoLife, GP: Grab-Posisi.}
\vspace{-2mm}
\label{tab:subset}
\end{table}

\par
\begin{figure}[h]
\centering    
  \vspace{-2mm}
  \includegraphics[width=0.85\linewidth]{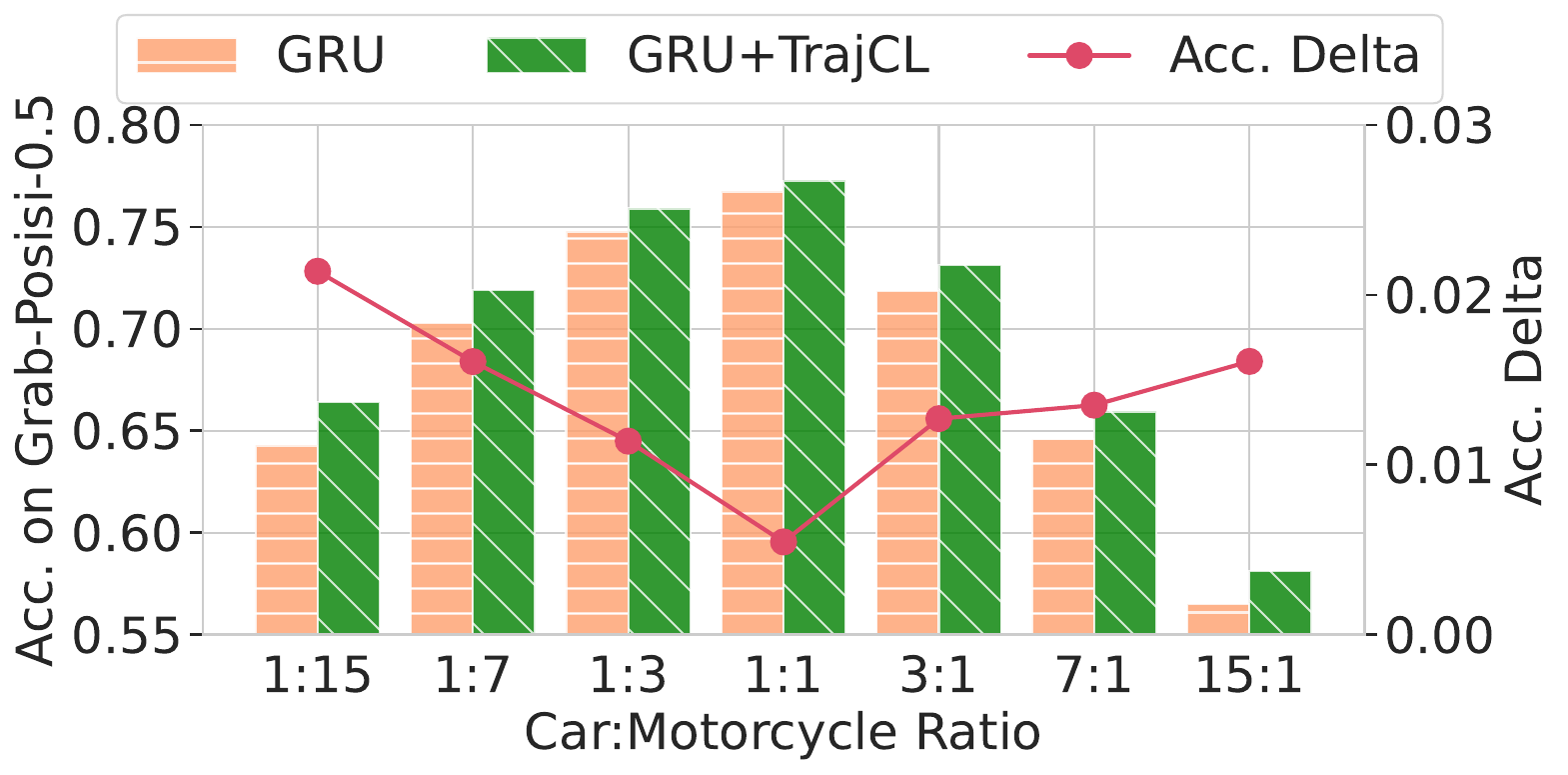}
    \vspace{-3mm}
  \caption{Exploration of imbalanced sample learning scenarios.}
  \vspace{-2mm}
  \label{fig:imbalance}
\end{figure}

 \noindent \textbf{Imbalanced Sample Learning.}
The results of imbalanced sample scenarios are summarized in Figure~\ref{fig:imbalance}, where the X-axis represents the Car:Motorcycle ratio, and the two Y-axes indicate the accuracy and performance improvement, respectively.
To focus on examining the effects of imbalanced samples and isolating the effects of other sophisticated design modules.
The Grab-Posisi dataset and GRU model are chosen here because this dataset only delineates car and motorcycle travel modes, and GRU is a relatively straightforward architecture for sequential modeling.

As depicted in Figure~\ref{fig:imbalance}, GRU+\model is consistently higher than the baseline, showing the effectiveness of the \model framework across all settings of imbalance.
As imbalances increase (from the middle to both sides), the red line in the figure illustrates more significant \model enhancements. 
The result further validates the robustness of \model, especially as it effectively addresses the challenges under severe imbalance conditions.
Interestingly, the performance gains are typically greater when there are fewer Car modes (left) than when Motorcycle is more prevalent (right). 
This can be intuitively attributed to the richer environmental conditions that motorcycles can encounter.
This complexity introduces a higher confounding effect, which in turn is elegantly addressed by our \model.

\subsection{Ablation Study (RQ3)}
\paragraph{Effects of Core Components.}
In the ablation study, we quantified the contribution of each \model component by removing the individual components.
Across both datasets, the results in Table~\ref{tab:abstudy}, demonstrate these performance degradations confirming the necessity of each component in the \model framework.
Replacing the environment codebook with a random matrix (w/o EC) slightly impacts performance, highlighting the learned prototypes for specific cities that can aid in better alignment.
Maintaining a fixed confounding degree of 0.5 for uniform disentanglement (w/o Dise) significantly affects performance, showing that soft-masks are crucial in separating confounders.
Similarly, excluding random combinations from the causal intervention process (w/o CI) leads to a similar decrease in performance, highlighting its importance in improving trajectory representation.
The most significant decrease occurs when geospatial features are excluded and only the raw trajectory is used to extract the confounding degree (\model w/o Env). This highlights the critical role of auxiliary environmental information in providing context for disentangling. 
These results demonstrate that proposed components collectively contribute to the effectiveness of \model. 
At the same time, geospatial features provide essential information for disentanglement, particularly evident in complex urban environments like the GeoLife dataset.

\begin{table}[t]
\setlength{\tabcolsep}{9pt}
\small
\centering
\begin{tabular}{lcccc} 
\toprule
\multirow{2}{*}{\diagbox[width=6em]{Variant}{Dataset}} & \multicolumn{2}{c}{GeoLife} & \multicolumn{2}{c}{Grab-Posisi} \\
\cmidrule(lr){2-3}\cmidrule(lr){4-5}
 & Acc.  &  $\Delta$  &  Acc.  &   $\Delta$ \\ 
\midrule
\model                  & 77.74  & --     & 79.47  & --            \\
\model w/o EC       & 76.25    & -1.49     & 76.99    & -2.48    \\
\model w/o CI       & 75.69    &  -2.05    & 76.39    & -3.08    \\
\model w/o Dise     & 75.55    & -2.19     & 76.23    & -3.24    \\
\rowcolor{gray!15}
\model w/o Env     & 70.16    & -7.58    & 65.38    & -14.09   \\
\bottomrule
\end{tabular}
\vspace{-2mm}
\caption{Component ablation results. We use GRU as backbone.}
\vspace{-4mm}
\label{tab:abstudy}
\end{table}

\begin{figure}[h]
  \vspace{-2mm}
  \includegraphics[width=1.0\linewidth]{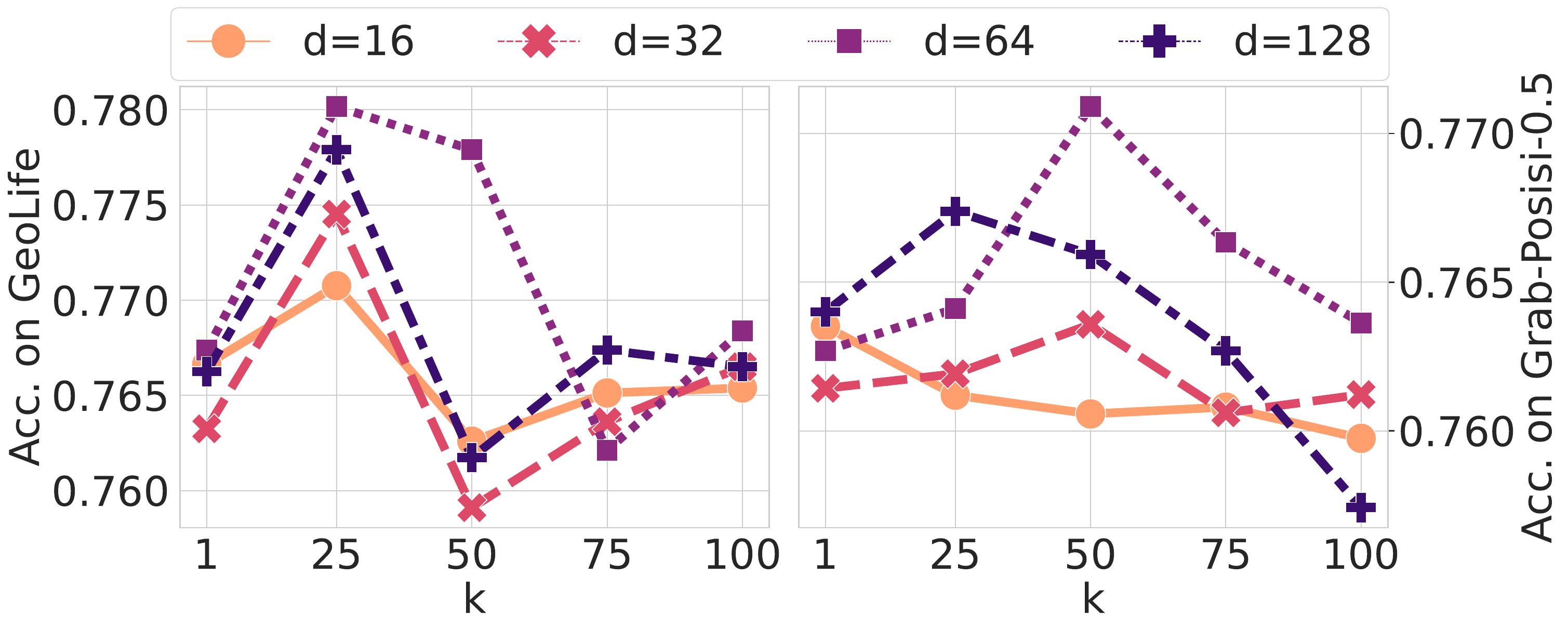}
  \vspace{-6mm}
  \caption{Effects of environment codebook size $k$ and hidden size $d$.}
  \vspace{-2mm}
  \label{fig:parameter}
\end{figure}

\paragraph{Effects of Environment Codebook.}
As discussed in Section~\ref{sec:env_module}, the environment codebook uses a learnable matrix to profile environmental prototypes.
Therefore, we explore the effect of the codebook parameters, analyzing the interaction between the richness of the environment (denoted by $k$) and the hidden dimension of each prototype (denoted by $d$).
As shown in Figure ~\ref{fig:parameter}, varying values of $k$ and $d$ lead to overall performance differences.
Each dataset has an optimal $k$ ($k=25$ in GeoLife and $k=50$ in Grab-Posisi).
In addition, the largest hidden dimension ($d=128$) does not necessarily yield the highest accuracy.
The reason is that a too large $d$ causes redundancy, while a small $d$ is insufficient to learn environmental features adequately.
We also find that the Geolife dataset is more sensitive to the codebook setting, which is also caused by the diversity of the environment.
This suggests that we need to balance $k$ and $d$ based on the specific urban forms and complexity.


\subsection{Interpretation Analysis (RQ4)}
\begin{figure}[t]
\centering    
  \includegraphics[width=0.95\linewidth]{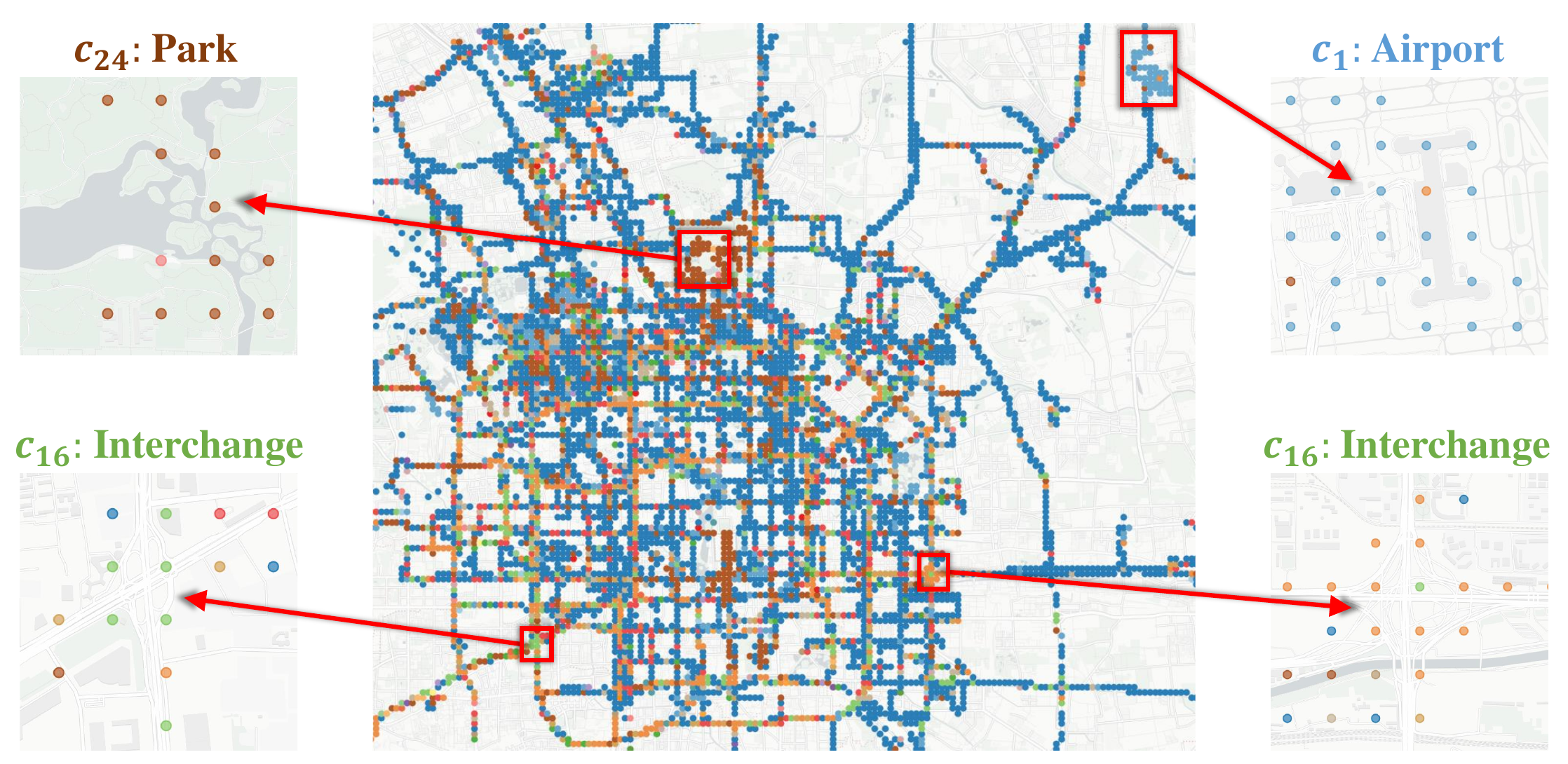}
  \caption{Visualization of environment codebook alignment on GeoLife dataset ($k$=25). The color represents different $c_i$.}
  \label{fig:codebook}
\end{figure}

To further explore what the codebook has learned, we plot trajectory points and their associated environmental prototypes to maps, analyzing the interpretability through geo-visualization.
As shown in Figure~\ref{fig:codebook}, we can clearly observe that the environment codebook can perceive different geospatial contexts.
For example, the high concentration of \textcolor[RGB]{162, 107, 71}{brown} (corresponding to prototype $c_{24}$ in the codebook) portrays a typical park area. The \textcolor[RGB]{46,117,182}{blue} (corresponding to prototype $c_{1}$) identifies a real-world airport. Meanwhile, the codebook assigns the identical environmental prototype (\textcolor[RGB]{121,178,83}{green} area) to those with the same geospatial contexts, demonstrating its strong ability to describe and categorize various environmental features.
In conclusion, the codebook can provide explanations, and also help us delve into the environmental perception of the model and its corresponding impact.

\begin{figure}[t]
  \includegraphics[width=1.0\linewidth]{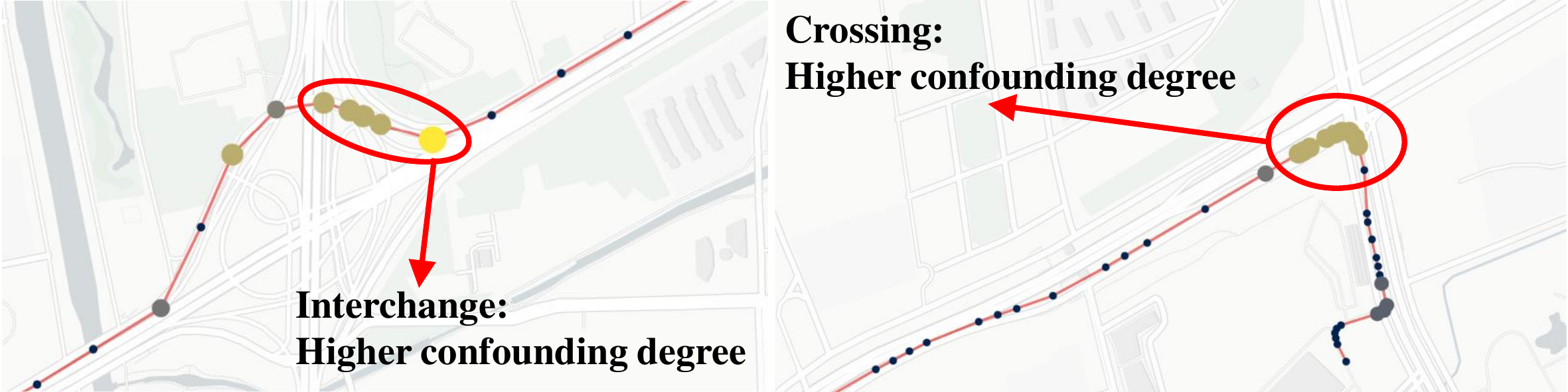}
  \caption{Visualization of the confounding soft-masks applied to two trajectories. Larger points indicate a higher confounding degree.}
  \label{fig:inter}
\end{figure}

Furthermore, we visualize the confounding soft-mask obtained through the environment alignment module for trajectory instances. 
This represents the confounding degree assigned for each trajectory point. 
As shown in Figure~\ref{fig:inter}, interchanges and crossings usually have higher degree. 
This aligns with our intuition to focus the confounding feature more on environmentally complex regions for effective decoupling.


\section{Conclusion and Future Work}

In this paper, we propose a novel causal trajectory modeling framework (\model) to facilitate the learning of robust and high-quality trajectory representations in travel mode identification task. \model reexamines existing trajectory modeling processes from a causal perspective, introducing an environment alignment module and a causal learning module for invariant trajectory representation learning. Experimental results on two real-world trajectory datasets show significant advantages in performance, robustness, and interpretability. 
In the future, we are committed to expanding the applicability of the 
\model to cover more diverse environments, thus enhancing its applicability to more realistic trajectory-based tasks such as travel time estimation.





\clearpage

\section*{Acknowledgments}
This work is mainly funded by Guangzhou-HKUST (GZ) Joint Funding Program (No. 2024A03J0620). This work is also supported by the Guangzhou Industrial Information and Intelligent Key Laboratory Project (No. 2024A03J0628).

\bibliographystyle{named}
\bibliography{ijcai24}

\clearpage
\appendix

\section{Appendix}

\subsection{Dataset Preprocessing and Statistics}\label{app:preprocess}

\subsubsection{Trajectory Data Processing}
When dealing with trajectory data, we encounter a significant variance in the number of points per trajectory and often face the challenge of limited labeling. For example, only about 2,000 labeled trajectories are available in the GeoLife dataset. Hence, the raw trajectories are first divided into shorter sub-trajectory instances using Algorithm \ref{algo}. This process can be seen as data augmentation.


\vspace{-2mm}
\begin{algorithm}[!h]
\SetAlgoLined
\KwData{$\{T_i\}^{N_T}_{i=1}$: Raw trajectories with labels.}
\KwIn{$N_{max}$: Maximum number of points in an instance;\\
~~~~~~~~~~~~$N_{min}$: Minimum number of points in an instance;\\
~~~~~~~~~~~~$M_{max}$: Maximum time range of an instance;\\
~~~~~~~~~~~~$M_{min}$: Minimum time range of an instance.}
\KwResult{$X$: Preprocessed trajectory instances.}
\BlankLine
$X \gets \{\emptyset\}$ \tcp*{Initialize instances set}
\For{$i \gets 1$ \KwTo $N_T$}{
    $M_i \gets$ \textup{Get the time range from} $T_i$ ; \\\hspace*{-2mm}
    \uIf{$M_i > M_{max}$}{
        $P \gets$ Partition $T_i$ into $M_{max}$-minute segments ; \\\hspace*{-2mm}
        \For{$j \gets 1$ \KwTo $|P|$}{
            $N_p \gets$ \textup{Get the number of points in} $P_j$ ; \\\hspace*{-2mm}
            \uIf{$N_p \geq N_{min}$ \textup{\textbf{and}} $N_p \leq N_{max}$}{
                Add $P_j$ to $X$ ;
            }
            \uElseIf{$N_p > N_{max}$}{
                $P'_j \gets$ Sample $N_{max}$ points from $P_j$ ;\\\hspace*{-2mm}
                Add $P'_j$ to $X$ ;
            }
        }
    }
    \uElseIf{\textup{get\_time\_range}($T_i$) $> M_{min}$}{
        Add $T_i$ to $X$\;
    }
}
\caption{Trajectory partition}
\label{algo}
\end{algorithm}

\vspace{-1mm}

To verify the generality of the \model and highlight its superior robustness, we increase the difficulty of the task, which aims at obtaining a good trajectory representation with a smaller number of points. Therefore, we set $N_{min}=20$ and $N_{max}=50$ on both datasets. Considering the differences in GPS sampling interval, for the GeoLife dataset we set $M_{min}=2$ and $M_{max}=10$; for the Grab-Posisi dataset, we set $M_{min}=2$ and $M_{max}=30$. Table \ref{app:dataset} provides a detailed overview of these preprocessed trajectory datasets.

\begin{table}[ht]
\setlength{\tabcolsep}{2pt}
\centering
\begin{tabular}{lcc}
\hline
\textbf{Attribute} & \textbf{GeoLife} & \textbf{Grab-Posisi} \\ \hline
\# instances       & 26,509           & 507,522              \\
\# points (range)  & 20--50           & 20--50               \\
\# points (avg)    & 48.7             & 46.0                   \\
\cmidrule(lr){1-3}
Duration (range)         & 2--10 minutes     & 2--30 minutes         \\
Duration (avg)           &  4.8 minutes     & 10.0 minutes         \\
Start date               & 2007/04/01       & 2019/04/10           \\
End date           & 2012/08/31       & 2019/04/16           \\
\cmidrule(lr){1-3}
\multirow{4}{*}{Class ratio} & Walk: 31.37\%      & Motorcycle: 49.92\%  \\
                            & Bike: 16.46\%      & Car: 50.08\%   \\
                            & Bus: 31.13\%       & -      \\
                             & Car: 21.04\%       & -         \\
\cmidrule(lr){1-3}
\multirow{2}{*}{MBR}          & 39.68°N--40.13°N  & 5.93°S--6.84°S        \\
         & 116.09°E--116.60°E& 105.88°E--107.41°E    \\ \hline
\end{tabular}
\caption{Data Summary for GeoLife and Grab-Posisi Datasets. MBR denotes the minimum bounding rectangle.}
\label{app:dataset}
\vspace{-1em}
\end{table}

\subsubsection{Geospatial Context Processing}
For the geospatial context, we extract the adjacent geospatial features of each trajectory point and arrange them into a sequence of the same length as the corresponding trajectory. To describe the environment in a universally applicable manner, we avoid complex data preprocessing.

First, we divide the geographic minimum bounding rectangle (MBR) into $200\times200$ square meters grids for two datasets. 
Subsequently, we retrieve 24 geospatial elements ($m=24$) from OpenStreetMap.com for each grid cell.
We intuitively count the occurrences of each geospatial element within the grid cell to convert these elements into numerical features. 
For instance, the number of traffic lights is a specific feature. 
Finally, we match the GPS point $p_i$ to its corresponding grid cell and retrieve 24 numerical features $e_i^1$ to $e_i^{24}$. 
The numerical features of each point in a trajectory are subsequently arranged into an environment sequence $E$.
Table \ref{app:envsum} presents an overview of 24 geospatial elements and their corresponding numerical features.

\begin{table}[!h]
\setlength{\tabcolsep}{5.5pt}
\centering
\begin{tabular}{ll cc cc}
\hline
\multirow{2}{*}{\textbf{Category}} 
& \multirow{2}{*}{\textbf{Element}} 

& \multicolumn{2}{c}{\textbf{GeoLife}} 
& \multicolumn{2}{c}{\textbf{Grab-Posisi}} \\
\cmidrule(lr){3-4}\cmidrule(lr){5-6}
& & Mean & Max& Mean & Max\\
\hline
\multirow{5}{*}{Traffic} 
& signals & 0.95 & 16.0 & 0.18 & 12.0 \\
& crossing & 0.98 & 26.0 & 0.39 & 16.0 \\
& junction & 0.09 & 6.0 & 0.03 & 2.0 \\
& parking & 0.29 & 11.0 & 0.17 & 11.0 \\
& fuel & 0.04 & 2.0 & 0.08 & 4.0 \\

\cmidrule(lr){1-6}
\multirow{3}{*}{Transport} 
& bus & 0.93 & 16.0 & 0.31 & 6.0 \\
& railway & 0.05 & 2.0 & 0.02 & 2.0 \\
& airport & 0.00 & 1.0 & 0.01 & 2.0 \\

\cmidrule(lr){1-6}
\multirow{6}{*}{Land use} 
& forest & 0.26 & 9.0 & 0.15 & 15.0 \\
& park & 0.35 & 12.0 & 0.14 & 9.0 \\
& retail & 0.06 & 9.0 & 0.05 & 42.0 \\
& residential & 1.62 & 19.0 & 6.45 & 42.0 \\
& commercial & 0.21 & 4.0 & 0.52 & 15.0 \\
& industrial & 0.02 & 4.0 & 0.15 & 6.0 \\

\cmidrule(lr){1-6}
\multirow{10}{*}{Road type} 
& trunk & 0.70 & 17.0 & 0.94 & 25.0 \\
& primary & 0.98 & 16.0 & 2.13 & 31.0 \\
& secondary & 0.87 & 20.0 & 0.95 & 22.0 \\
& tertiary & 2.14 & 18.0 & 1.13 & 24.0 \\
& footway & 3.30 & 87.0 & 0.67 & 43.0 \\
& link & 1.16 & 22.0 & 1.13 & 27.0 \\
& cycleway & 0.86 & 16.0 & 0.02 & 21.0 \\
& motorway & 1.21 & 13.0 & 0.95 & 15.0 \\
& service & 1.73 & 38.0 & 7.30 & 60.0 \\
& steps & 0.58 & 21.0 & 0.29 & 11.0 \\
\hline
\end{tabular}

\caption{Description of geospatial context data. The category and element are sourced from OpenStreetMap. The numerical values indicate the count of geographical elements near each trajectory point.}
\vspace{-1em}
\label{app:envsum}
\end{table}

\subsubsection{Data Partition in Robustness Test}
The data partition details in the robustness test (RQ2) are presented below.

\begin{itemize}[leftmargin=*]

\item \textbf{Few-shot Learning}: 
To ensure a fair comparison, we only sample subsets from the original training set at ratios of 0.1, 0.2, and 0.5, while keeping the test/validation set intact.
\item \textbf{Imbalanced Sample Learning}: 
We keep the test and validation sets constant while preserving a balanced category distribution (1:1 car to motorcycle ratio in the Grab-Posisi dataset). 
The total amount of training data is kept consistent, equivalent to half of the original training set, with adjustments made solely to the category proportions.

\end{itemize}


\subsection{Experimental Settings}\label{app:exp_setting}

\subsubsection{More Description and Settings of Baselines}

We implement our \model framework with five representative baselines as follows:

\begin{itemize}[leftmargin=*]

\item \textbf{GRU}~\cite{cho2014learning}: A vanilla gated recurrent unit network applied for general sequence modeling tasks.

\item \textbf{BiLSTM}~\cite{liu2017end}: A traditional framework for classifying trajectory travel modes that utilize an end-to-end bidirectional LSTM model, integrating time interval embeddings as additional external features.

\item \textbf{GRU-D}~\cite{che2018recurrent}: A modified GRU model that introduces a novel time gate designed to capture the time irregularity by applying a time decay term in sequences.

\item \textbf{STGN}~\cite{zhao2020go}: An LSTM variant incorporating spatial and temporal gating mechanisms to better understand the spatio-temporal intervals between nearby points.

\item \textbf{TrajFormer}~\cite{liang2022trajformer}: The state-of-the-art method adapts transformers to model trajectories. It generates continuous point embedding to jointly consider input features and spatio-temporal intervals. A squeeze function is then used to speed up representation learning. 

\end{itemize}

For a fair comparison, each baseline is constructed as a sequence encoder equivalent to the single-branch form of the \model.
Specifically, two $3\times1$ convolutional layers are uniformly applied before the sequence encoder to expand the channel. After the sequence encoder, a 2-layer MLP is used as the predictor.
Additionally, the direct input for all baselines consistently includes each GPS point ${p}= \left({Lon}, {Lat}, t\right)$ and its surrounding environment $E$. Hence, we only fuse an external feature $distance$ at the top layer of STGN and the continuous point embedding block of TrajFormer.

\subsubsection{More Implementation Details of \model}
Our \model framework is implemented with PyTorch 2.1.2 and all experiments are conducted on Nvidia RTX 3090. The default embedding dimension is set to 64, covering the kernel size of convolutional layers, the hidden dimensions of both confounding and causal encoders, and the output dimension of the first linear layer in the predictors. For the environment codebook in our comparative experiments, the hidden dimension for each prototype $d$ equals 64, and the codebook size $k$ is 50. We conduct a grid search for the weight parameters and set $\lambda$, $\varphi$, and $\eta$ to 1, 0.5, and 0.5, respectively.


\subsection{More Details on Experimental Results}

\subsubsection{More Results of Imbalanced Sample Learning}

To thoroughly verify the performance in the imbalanced sample learning scenario, we further implement \model to two baselines, GRU-D and Trajformer, under identical settings. 
Table \ref{app:imbal} illustrates that \model steadily improves performance across different backbones and category ratios.
In addition, the improvements of \model are more significant as the imbalance ratio escalates, which aligns with our previous findings and confirms the robustness of our framework.

\begin{table}[ht]
\vspace{2mm}
\small
\setlength{\tabcolsep}{3.5pt}
\centering
\begin{tabular}{c ccccccc}
\hline

\toprule
\multirow{2}{*}{Variant} & \multicolumn{7}{c}{Car:Motorcycle Ratio} \\
\cmidrule(lr){2-8}

& 15:1 & 7:1 & 3:1 & 1:1 & 1:3 & 1:7 & 1:15 \\
\midrule

GRU-D & 56.92 & 63.51 & 71.47 & 75.82 & 73.64 & 68.52 & 64.13 \\
+~\model & 58.22 & 64.69 & 72.28 & 76.67 & 74.51 & 69.37 & 65.25 \\
\rowcolor{gray!15}
$\Delta$ & +1.30 & +1.18 & +0.81 & +0.85 & +0.87 & +0.85 & +1.12 \\

\cmidrule(lr){1-8}

TrajFormer & 58.86 & 66.31 & 74.92 & 79.97 & 77.19 & 71.58 & 65.61 \\
+~\model & 60.12 & 67.46 & 76.10 & 81.12 & 78.28 & 72.75 & 66.83 \\
\rowcolor{gray!15}
$\Delta$ & +1.26 & +1.15 & +1.18 & +1.15 & +1.09 & +1.17 & +1.22 \\
\hline
\end{tabular}
\caption{Performance in the imbalanced sample learning on other backbones. Accuracy is reported by percentage (\%)}
\label{app:imbal}
\vspace{-1em}
\end{table}

\subsubsection{More Interpretation Analysis}

\begin{figure}[!b]
  \includegraphics[width=1.0\linewidth]{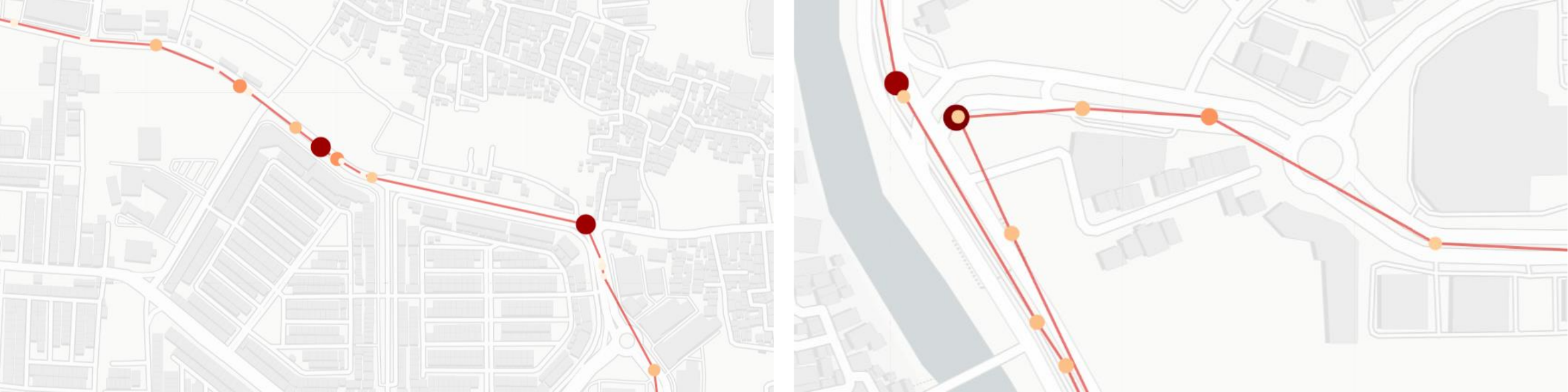}
  \par\vspace{3mm}
  \includegraphics[width=1.0\linewidth]{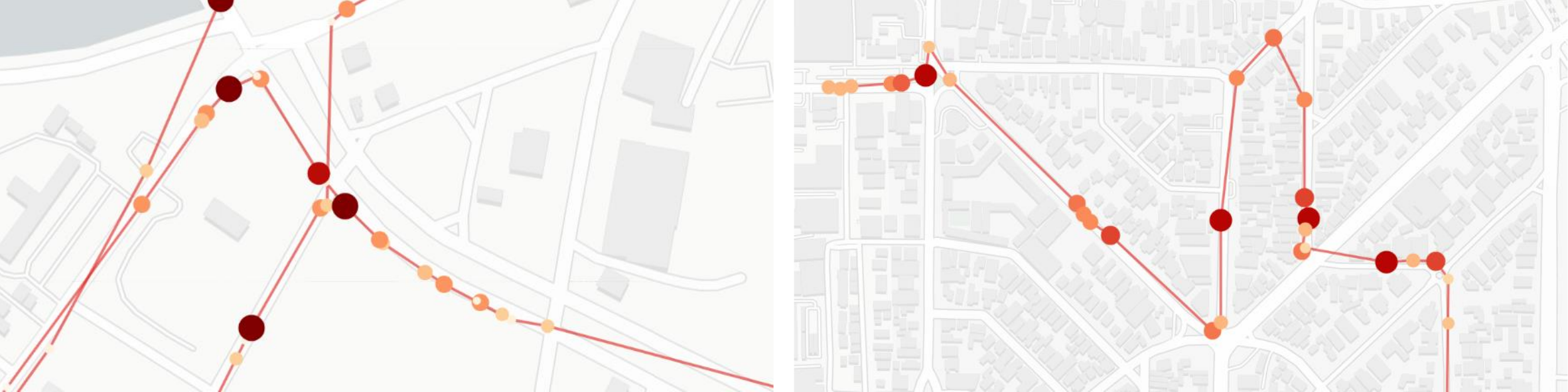}
  \par\vspace{3mm}
  \includegraphics[width=1.0\linewidth]{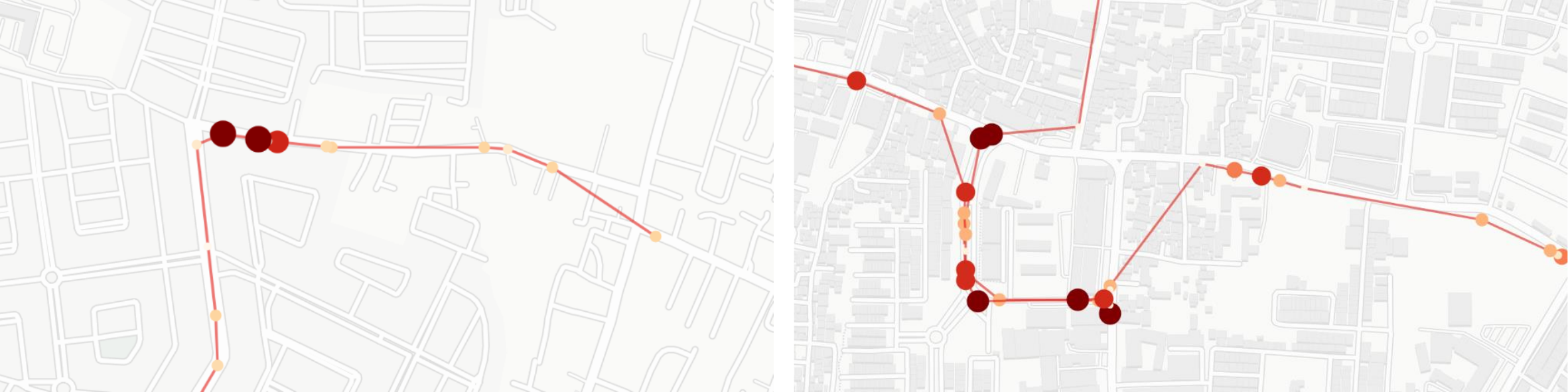}
  \caption{Visualization of the confounding soft-masks applied to trajectories in the Grab-Posisi dataset. Larger points indicate a higher confounding degree caused by the surrounding geospatial context.}
  \label{app:grab}
\end{figure}

\begin{figure*}[h]
  \includegraphics[width=1.0\linewidth]{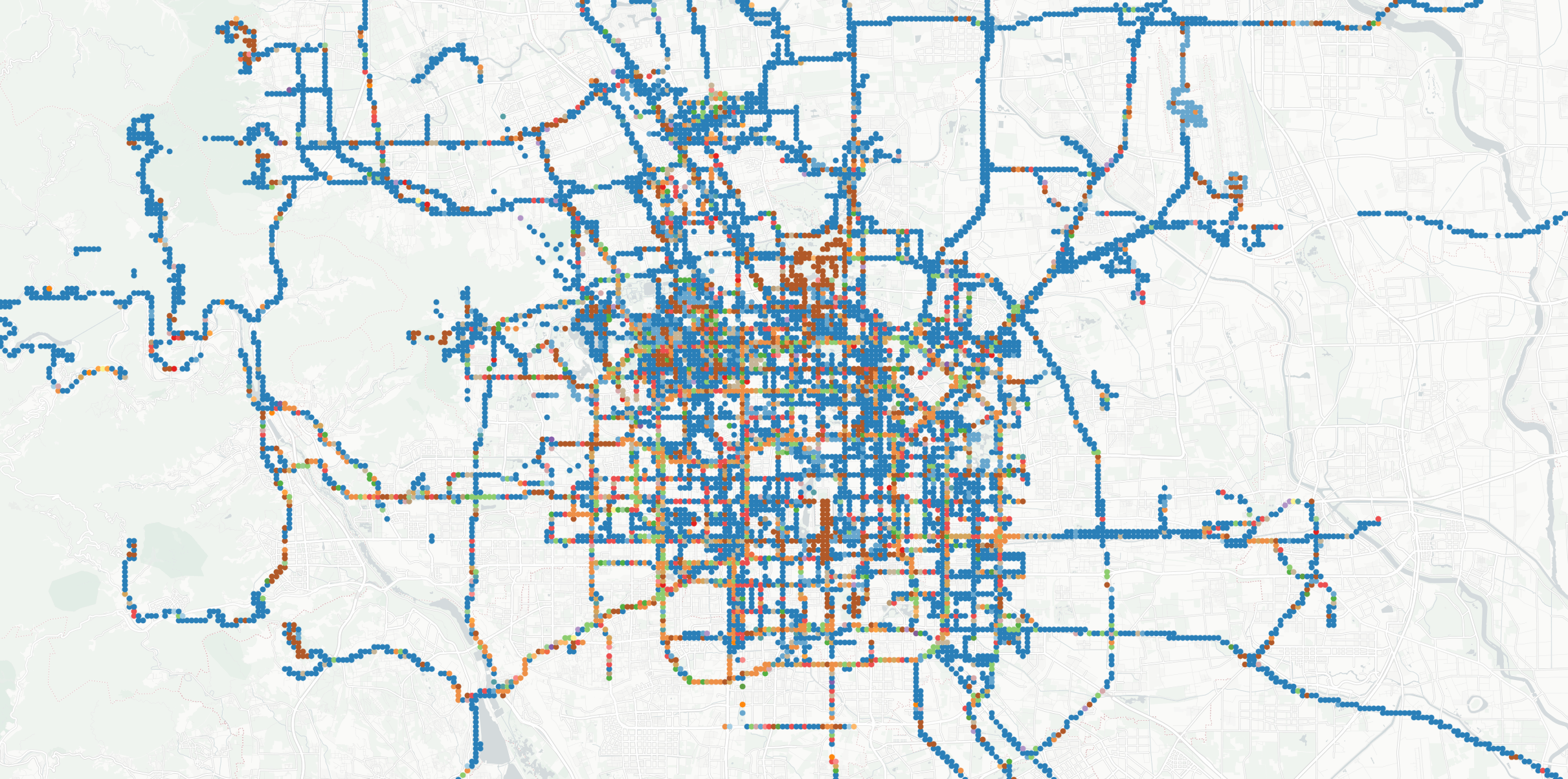}
  \par\vspace{3mm}
  \caption{Visualization of the environment codebook aligned with the Beijing map using the GeoLife dataset ($k$=25). The color represents different $c_i$. 
  We found that \textcolor[RGB]{237,82,83}{$\boldsymbol{c_{10}}$} represents road sections prone to congestion, \textcolor[RGB]{177,89,40}{$\boldsymbol{c_{24}}$} represents typical parks and recreational areas, \textcolor[RGB]{148,201,108}{$\boldsymbol{c_{5}}$} represents interchange areas, and \textcolor[RGB]{199,182,153}{$\boldsymbol{c_{21}}$} represents the typical residential area. This demonstrates that the environment codebook can effectively distinguish and discretize different geospatial contexts to assist in decoupling.}
  \label{app:bjmapfig}
\end{figure*}

\begin{figure*}[h]
  \includegraphics[width=1.0\linewidth]{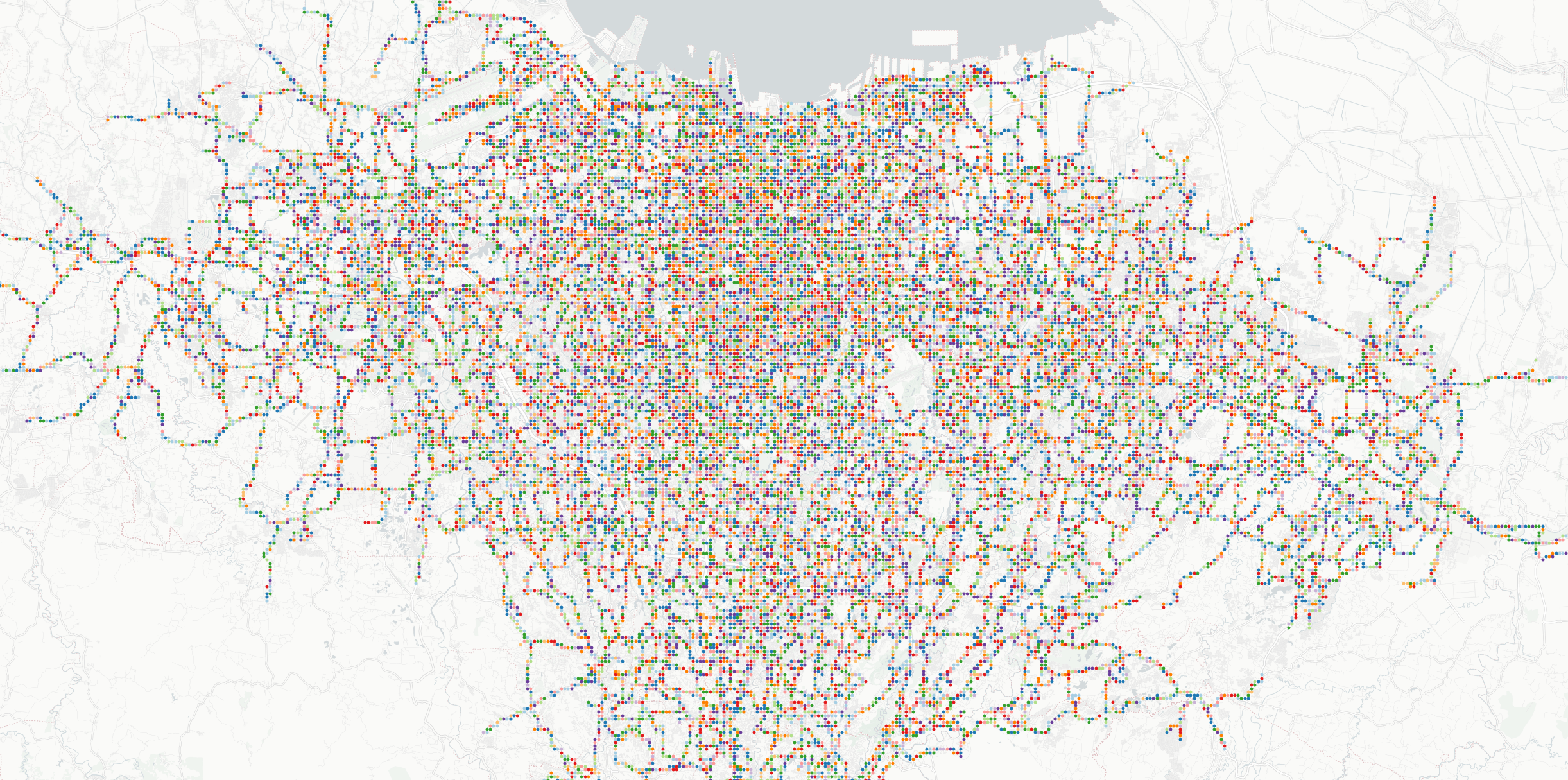}
  \par\vspace{3mm}
  \caption{Visualization of the environment codebook aligned with the Jakarta map using the Grab-Posisi dataset ($k$=10). The color represents different $c_i$. This dataset contains a larger number of trajectories and covers a broader area. Downtown areas have a more diverse environment. Our analysis also indicates that the environment codebook usually assigns only a few prototypes to suburban areas, while downtown areas have multiple prototypes.}
  
  \label{app:jakmapfig}
\end{figure*}

In this subsection, we provide additional interpretative analysis.
Figure~\ref{app:grab} presents the visualizations of the confounding soft-masks applied to more trajectories in the Grab-Posisi dataset. 
Figures~\ref{app:bjmapfig} and~\ref{app:jakmapfig} display the expanded visualizations of the environmental codebook aligned with the map.

\end{document}